\documentclass[runningheads]{llncs}
\usepackage[T1]{fontenc}
\usepackage{graphicx}
\usepackage{tikz}
\usetikzlibrary{positioning, fit}
\usepackage{longtable}
\usepackage{booktabs}    
\usepackage{multirow}    
\usepackage{graphicx}    
\usepackage{tabularx}    
\usepackage{rotating}
\usepackage{makecell}
\usepackage{amsmath}
\usepackage[linesnumbered,ruled,vlined]{algorithm2e}
\usepackage{hyperref}
\usepackage{listings}
\usepackage[most]{tcolorbox}
\usepackage{pgfplots}
\usepackage{enumitem}
\pgfplotsset{compat=1.18}

\usepackage{xcolor}
\newcommand{\new}[1]{#1}

\newtcolorbox{promptbox}[1][]{
  colback=white,        
  colframe=black!30,    
  fonttitle=\bfseries,
  title=Prompt: #1,
  sharp corners,
  boxrule=0.4pt,        
  arc=1pt,
  left=6pt,
  right=6pt,
  top=4pt,
  bottom=4pt,
  enhanced,
}

\begin{document}
%
\title{Semantic-Guided Reading Order Reconstruction in Historical Armenian Newspapers with LLMs}
\titlerunning{Complex Reading Order in Armenian}
%
\author{Chahan Vidal-Gorène\inst{1,2,4}\orcidID{0000-0003-1567-6508} \and
Nadi Tomeh\inst{2}\orcidID{0000-0002-1813-3579} \and
Victoria Khurshudyan\inst{3}\orcidID{0000-0002-9960-2516}}
%
\authorrunning{Vidal-Gorène et al.}
%
\institute{École nationale des chartes-PSL \and
Laboratoire d'Informatique Paris Nord \and
Inalco \and
Calfa\\
\email{chahan.vidal-gorene@chartes.psl.eu}}
%
\maketitle              
\begin{abstract}
This paper addresses reading order reconstruction in historical Armenian newspapers, which combine complex layouts with limited language resources. We introduce a new annotated dataset of 66 pages and compare geometric heuristics, YOLO-based layout parsing, an end-to-end document model ECLAIR, and a hybrid method combining semantic zone detection with a generative LLM. Our hybrid method achieves the lowest error rates of all evaluated approaches, reducing ordering errors by up to 76\% over the strongest geometric baseline, and remains robust in multi-page settings and under noisy OCR. Rather than targeting production the method is designed as a data bootstrapping strategy enabling rapid annotation in highly under-resourced scenarios. Alongside the dataset, we release a specialized Tesseract OCR model for historical Armenian print.
\end{abstract}

\section{Introduction}

The Armenian press spans over two centuries, from \textit{Azdarar} (Madras, 1794) through a 19th-century golden age in Constantinople, Tiflis, Moscow, and the early diaspora; the 1915 Genocide redirected publishing toward Middle Eastern, European, and American diasporas, while the Soviet era (1920--1991) and post-1991 independence each produced their own periodicals. The resulting corpus combines Classical, Western, and Eastern Armenian (with frequent multilingual inserts), evolving orthographies, and printing technologies ranging from letterpress and lithography to offset and clandestine mimeographs—yielding highly heterogeneous scan qualities. Our work targets OCR and structuring of thousands of these pages, mostly in Western Armenian\footnote{Western Armenian, traditionally spoken by Ottoman Armenians and now mostly diasporic, is one of the two standardized varieties of Modern Armenian. It differs from Eastern Armenian in phonology, vocabulary, and certain grammatical features, and uses classical orthography. UNESCO classifies it as endangered.}, contributing to a broader effort to build a Western Armenian reference corpus and to enable downstream resources for an endangered language.

A major obstacle is accurate reading order reconstruction: these newspapers use multi-column designs with irregular article placement, overlapping titles, and non-linear flows, and OCR models adapted to the script still produce disordered outputs. We address this by introducing a newly annotated dataset of Armenian newspapers, proposing a specialized OCR model for historical Armenian, and comparing strategies—layout heuristics, YOLO-based detectors, and large language models—under realistic OCR constraints, both within and across pages.

Our LLM-based methods are framed as an \emph{annotation accelerator}—generating candidate orders that humans verify—rather than a production pipeline, since cloud-API LLMs scale poorly in cost and latency.

\section{Related Works}\label{sec:related_works}

Reading order detection is a key component of document structure analysis, impacting information retrieval, OCR post-processing, and large-scaxle dataset preparation for LLMs. Traditional methods rely on heuristics (top-to-bottom, left-to-right) or classical models based on Bayesian inference and logic programming~\cite{breuel2003high}; both struggle with multi-column and irregular layouts. Deep learning brought layout-aware architectures: LayoutLM~\cite{xu2020layoutlm} integrates visual and positional cues with text, and LayoutReader~\cite{wang2021layoutreader} predicts reading order directly with a sequence-to-sequence paradigm. End-to-end document models such as Kosmos-2.5~\cite{lv2024kosmos25multimodalliteratemodel}, GOT~\cite{wei2024general}, Nougat~\cite{blecher2023nougatneuralopticalunderstanding}, and ECLAIR~\cite{karmanov2025eclair} combine OCR, region detection, and reading order in a unified encoder-decoder (ECLAIR uses a ViT-H (RADIO) encoder with an mBART decoder~\cite{liu2020multilingual} and was benchmarked against LayoutLMv3, Kosmos-2.5, and GOT). These systems excel on structured English corpora but perform poorly on degraded documents and minority scripts. Two-stage pipelines remain common: Quiros et al.~\cite{quiros2018multi} separate layout detection from sequence modeling, and recent work on Chinese historical documents~\cite{bizais2024optimizing} adopts a similar hierarchical strategy.

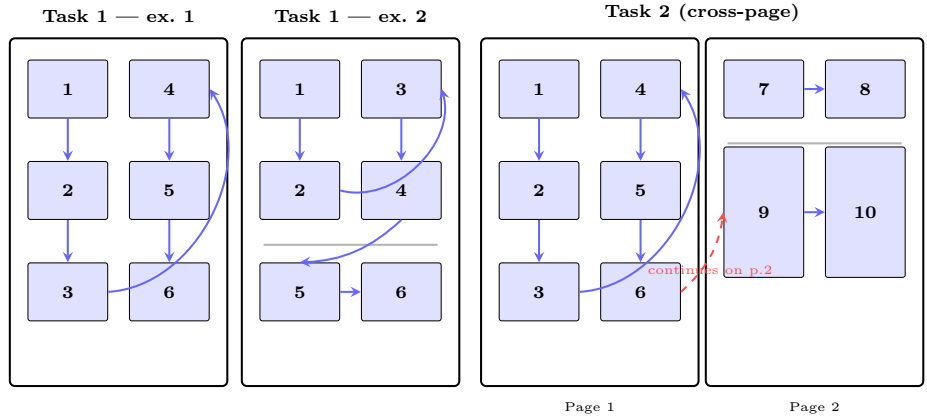
\begin{figure}[htbp]
\centering
{
\resizebox{\textwidth}{!}{%
\begin{tikzpicture}[
  font=\small,
  page/.style={draw, thick, rounded corners=2pt, minimum width=30mm, minimum height=48mm},
  zone/.style={draw, fill=blue!12, rounded corners=1pt, align=center, font=\scriptsize, minimum width=11mm, minimum height=8mm, inner sep=1pt},
  arr/.style={->, >=stealth, thick, dashed}
]

\node[page] (p1) at (0,0) {};
\node[above=1mm of p1, font=\scriptsize\bfseries] {Task 1 — ex.~1};

\node[zone] (z1) at (-0.7,  1.7)  {\textbf{1}};
\node[zone] (z2) at (-0.7,  0.3)  {\textbf{2}};
\node[zone] (z3) at (-0.7, -1.1)  {\textbf{3}};
\node[zone] (z4) at ( 0.7,  1.7)  {\textbf{4}};
\node[zone] (z5) at ( 0.7,  0.3)  {\textbf{5}};
\node[zone] (z6) at ( 0.7, -1.1)  {\textbf{6}};

\draw[->, >=stealth, blue!60, thick] (z1) -- (z2);
\draw[->, >=stealth, blue!60, thick] (z2) -- (z3);
\draw[->, >=stealth, blue!60, thick] (z3.east) to[bend right=60] (z4.east);
\draw[->, >=stealth, blue!60, thick] (z4) -- (z5);
\draw[->, >=stealth, blue!60, thick] (z5) -- (z6);

\node[page] (p2) at (3.2,0) {};
\node[above=1mm of p2, font=\scriptsize\bfseries] {Task 1 — ex.~2};

\node[zone] (w1) at (2.5,  1.7) {\textbf{1}};
\node[zone] (w2) at (2.5,  0.3) {\textbf{2}};
\node[zone] (w3) at (3.9,  1.7) {\textbf{3}};
\node[zone] (w4) at (3.9,  0.3) {\textbf{4}};
\draw[gray!60, thick] (2.0, -0.45) -- (4.4, -0.45);
\node[zone] (w5) at (2.5, -1.1) {\textbf{5}};
\node[zone] (w6) at (3.9, -1.1) {\textbf{6}};

\draw[->, >=stealth, blue!60, thick] (w1) -- (w2);
\draw[->, >=stealth, blue!60, thick] (w2.east) to[bend right=60] (w3.east);
\draw[->, >=stealth, blue!60, thick] (w3) -- (w4);
\draw[->, >=stealth, blue!60, thick] (w4.south) to[bend left=20] (w5.north);
\draw[->, >=stealth, blue!60, thick] (w5) -- (w6);

\node[page] (pa) at (6.5,0) {};
\node[page] (pb) at (9.6,0) {};
\node[above=1mm of pa, xshift=15.5mm, font=\scriptsize\bfseries] {Task 2 (cross-page)};

\node[zone] (a1)  at (5.8,  1.7)  {\textbf{1}};
\node[zone] (a2)  at (5.8,  0.3)  {\textbf{2}};
\node[zone] (a3)  at (5.8, -1.1)  {\textbf{3}};
\node[zone] (ac1) at (7.2,  1.7)  {\textbf{4}};
\node[zone] (ac2) at (7.2,  0.3)  {\textbf{5}};
\node[zone] (ac3) at (7.2, -1.1)  {\textbf{6}};

\node[zone] (b1) at (8.9,  1.7) {\textbf{7}};
\node[zone] (b2) at (10.3, 1.7) {\textbf{8}};
\draw[gray!60, thick] (8.4, 0.95) -- (10.8, 0.95);
\node[zone, minimum height=18mm] (b3) at (8.9, 0.0) {\textbf{9}};
\node[zone, minimum height=18mm] (b4) at (10.3, 0.0) {\textbf{10}};

\draw[->, >=stealth, blue!60, thick] (a1) -- (a2);
\draw[->, >=stealth, blue!60, thick] (a2) -- (a3);
\draw[->, >=stealth, blue!60, thick] (a3.east) to[bend right=60] (ac1.east);
\draw[->, >=stealth, blue!60, thick] (ac1) -- (ac2);
\draw[->, >=stealth, blue!60, thick] (ac2) -- (ac3);
\draw[arr, red!70] (ac3.east) to[bend right=20] node[below, font=\tiny] {continues on p.2} (b3.west);
\draw[->, >=stealth, blue!60, thick] (b1) -- (b2);
\draw[->, >=stealth, blue!60, thick] (b3) -- (b4);

\node[below=1mm of pa, font=\tiny] {Page 1};
\node[below=1mm of pb, font=\tiny] {Page 2};
\end{tikzpicture}
}}
\caption{\new{\textbf{Task~1, ex.~1}: two-column page, read column by column (zones~1--3 down column~1, then zones~4--6 down column~2). \textbf{Task~1, ex.~2}: two upper columns (zones~1--2, then~3--4) separated by a horizontal rule from a bottom row (zones~5--6 read left-to-right). \textbf{Task~2}: an article begins on page~1 (zones~1--6, column-major order) and continues on page~2 as zone~9, below a horizontal separator and unrelated zones~7--8 (dashed red arrow).}}
\label{fig:tasks}
\end{figure}

Sequence-to-sequence models such as LayoutReader and LayoutLM-based approaches require token-level training data in the target script; their fine-tuning for Western Armenian is precluded by the absence of large annotated corpora—precisely the under-resourced setting this work targets.

\section{Dataset}\label{sec:dataset}

Digital archives of the Armenian press—mostly scans rather than plain text—are accessible through four main institutions\footnote{Pan-Armenian Digital Library ARAR, maintained by the Fundamental Scientific Library of Armenia (FSL) [\href{https://arar.sci.am/}{arar.sci.am}], with notable holdings from the Mekhitarist Congregation Library in Vienna; the National Library of Armenia (NLA) [\href{https://tert.nla.am/}{tert.nla.am}], covering early 20th c.\ to post-Soviet periodicals; the Armenian Research and Archives Museum (ARAM) [\href{https://webaram.com/}{webaram.com}], focused on diaspora press; and BULAC [\href{https://www.bulac.fr/}{bulac.fr}], specializing in 20th c.\ diaspora publications.}. The scanned data used in the present study are drawn from one or more of these archives (for further details, see Git repository).

The dataset consists of 66 scanned pages (1913--2009) with high variability in layout, scan quality and typography, provided in \emph{PageXML} and \emph{YOLO} bounding-box formats. All annotations were created using the Calfa Vision annotation tool~\cite{vidal2021modular}. While the dataset primarily reflects Armenian newspaper production from the French diaspora throughout the 20\textsuperscript{th} century, the diversity of layouts, typefaces, and digitization conditions remains representative of broader Armenian newspaper design, excluding periodicals.

Three levels of annotation have been defined: (i) semantic text regions to describe document content, (ii) semantic areas used for reading order experiments, and (iii) structural separators. A total of 11 classes are annotated. The classes and their frequencies across the dataset are summarized in Table~\ref{tab:classes} and Table~\ref{tab:visual_examples}.

\begin{table}[ht]
    \centering
    \begin{tabularx}{\textwidth}{p{3cm}|p{1.5cm}|X}
        \textbf{Class} & \textbf{Total} & \textbf{Description} \\\hline
        Advertisement & 38 & Typically found on the last page; short text, address, often framed; not semantically linked to content. \\\hline
        HorizontalSeparator & 93 & Horizontal visual separator between layout elements. \\\hline
        VerticalSeparator & 91 & Vertical visual separator between layout elements. \\\hline
        Marginalia & 34 & Any text or annotation in the margins (e.g., stamps, handwritten notes). \\\hline
        Metadata & 87 & Header information on the front page (editor, address, price, etc.). \\\hline
        PageNumber & 38 & Page numbering elements. \\\hline
        Paragraph & 2380 & Text blocks starting with indentation; tables are not present; heavily indented text is treated as paragraphs. \\\hline
        RunningTitle & 50 & Repeated headers, typically at the top of pages. \\\hline
        SemanticBlock & 421 & Grouped content zone (e.g., an article spanning multiple regions). \\\hline
        Signature & 120 & Author name at the end of a column, usually right-aligned. \\\hline
        Title & 601 & Any title; a new bounding box is created for each change in font or font size. \\\hline
    \end{tabularx}
    \caption{Annotated classes in the dataset, including instance counts and descriptions.}
    \label{tab:classes}
\end{table}

The dataset includes pages from 35 distinct newspaper issues (1913--2009), primarily in Western Armenian (MWA)—the under-resourced variant we target—with a few examples in Eastern Armenian (MEA) and some bilingual MWA / French issues. Scan quality ranges from good to poor (blur, damage, curvature, uneven exposure), and layouts span compact book-like formats to wide newspaper pages with two to six columns. For each issue, we selected two pages: typically the front or last page plus one interior page containing content that continues from the front page (Tasks 1 and 2, see Fig.~\ref{fig:tasks}).

{\paragraph*{Dataset splits.}
For all reading order experiments, we use a stratified 80/10/10 train/val/test split, stratified by newspaper title to prevent leakage across issues of the same publication. An additional 20 out-of-domain images (unseen newspaper titles, sourced from the same digital libraries) form an independent test set for Tables~\ref{tab:ocr_results} and~\ref{tab:layout_analysis}; Tables~\ref{tab:task1_results} and~\ref{tab:task2_results} use the in-domain test split.}

\paragraph*{Synthetic data.}
We augment the corpus to 500 images via image-to-image translation~\cite{CycleGAN2017,vidal2024image} from semantic masks derived from PageXML annotations (see Table~\ref{tab:visual_examples}); the masks guide the generation of realistic newspaper-like layouts without textual content. To mimic typical scan degradations we apply randomized transformations from the \emph{Albumentations} library (\texttt{ElasticTransform}, \texttt{RandomBrightnessContrast}, \texttt{Flip+Flop}, \texttt{RandomFog}, \texttt{GaussianBlur}, \texttt{ZoomBlur}). Each high-resolution image (typically $>3000\times5000$\,px) is split into four quadrants prior to generation and recombined afterwards to preserve coherence.

\begin{table}[htbp]
    \centering
    \renewcommand{\arraystretch}{1.2}
    \resizebox{\textwidth}{!}{%
    \begin{tabular}{p{3cm} c c c}
        & \textbf{3 cols. (1949)} & \textbf{4 cols. (2001)} & \textbf{5 cols. (1937)} \\
        \textbf{Image} &
        \includegraphics[width=0.22\textwidth]{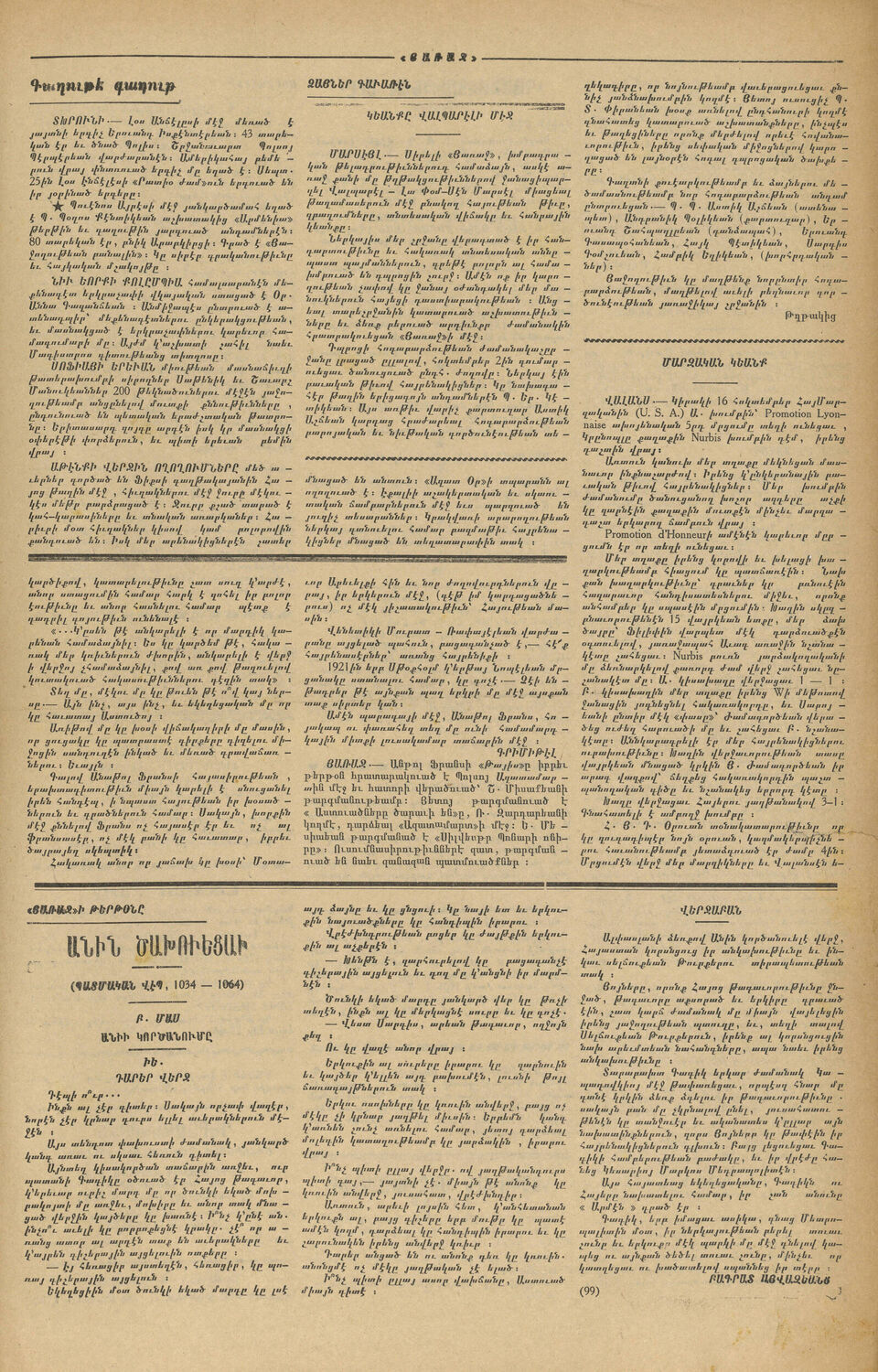} &
        \includegraphics[width=0.22\textwidth]{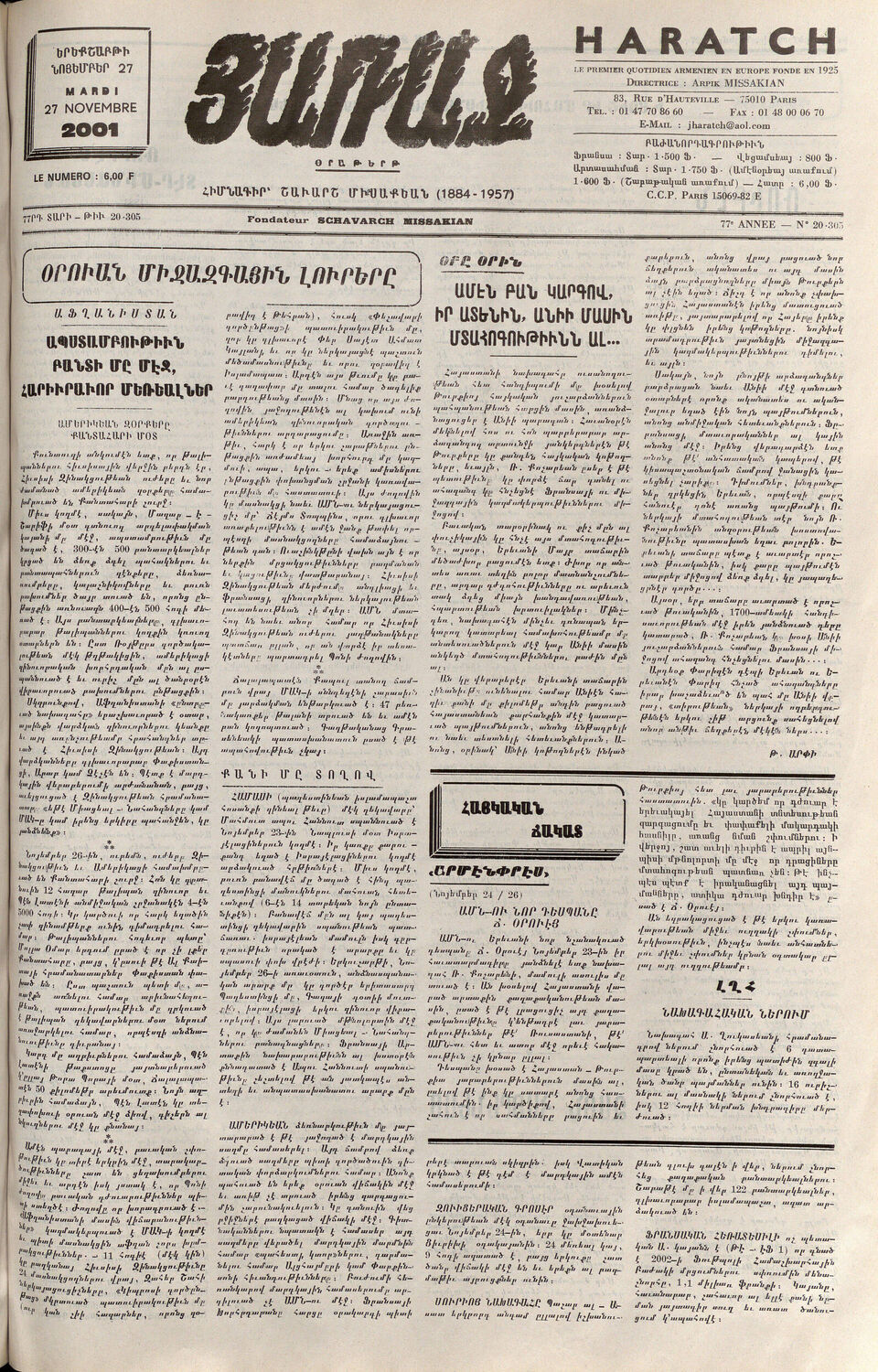} &
        \includegraphics[width=0.22\textwidth]{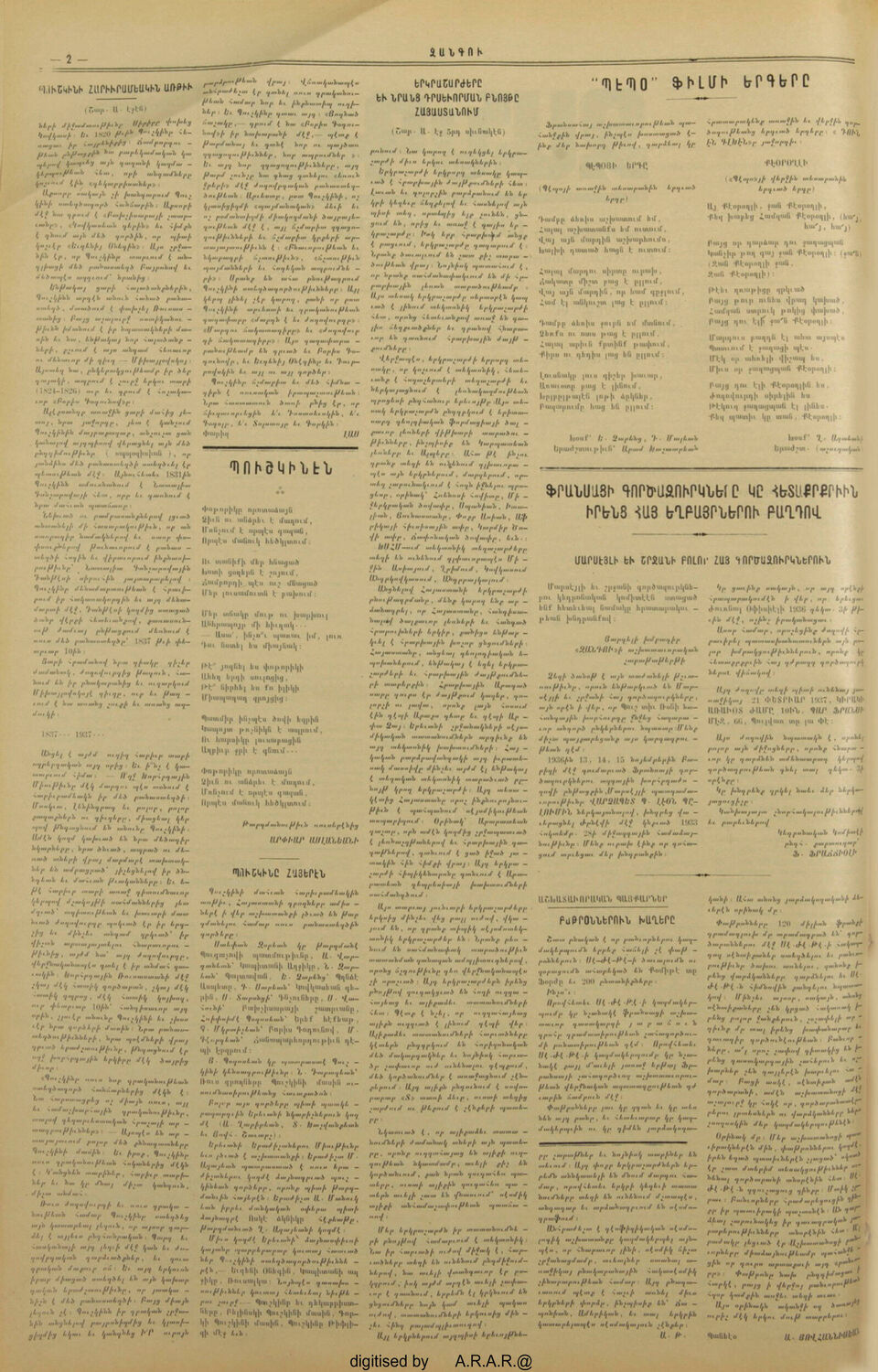} \\

        \textbf{Text regions} &
        \includegraphics[width=0.22\textwidth]{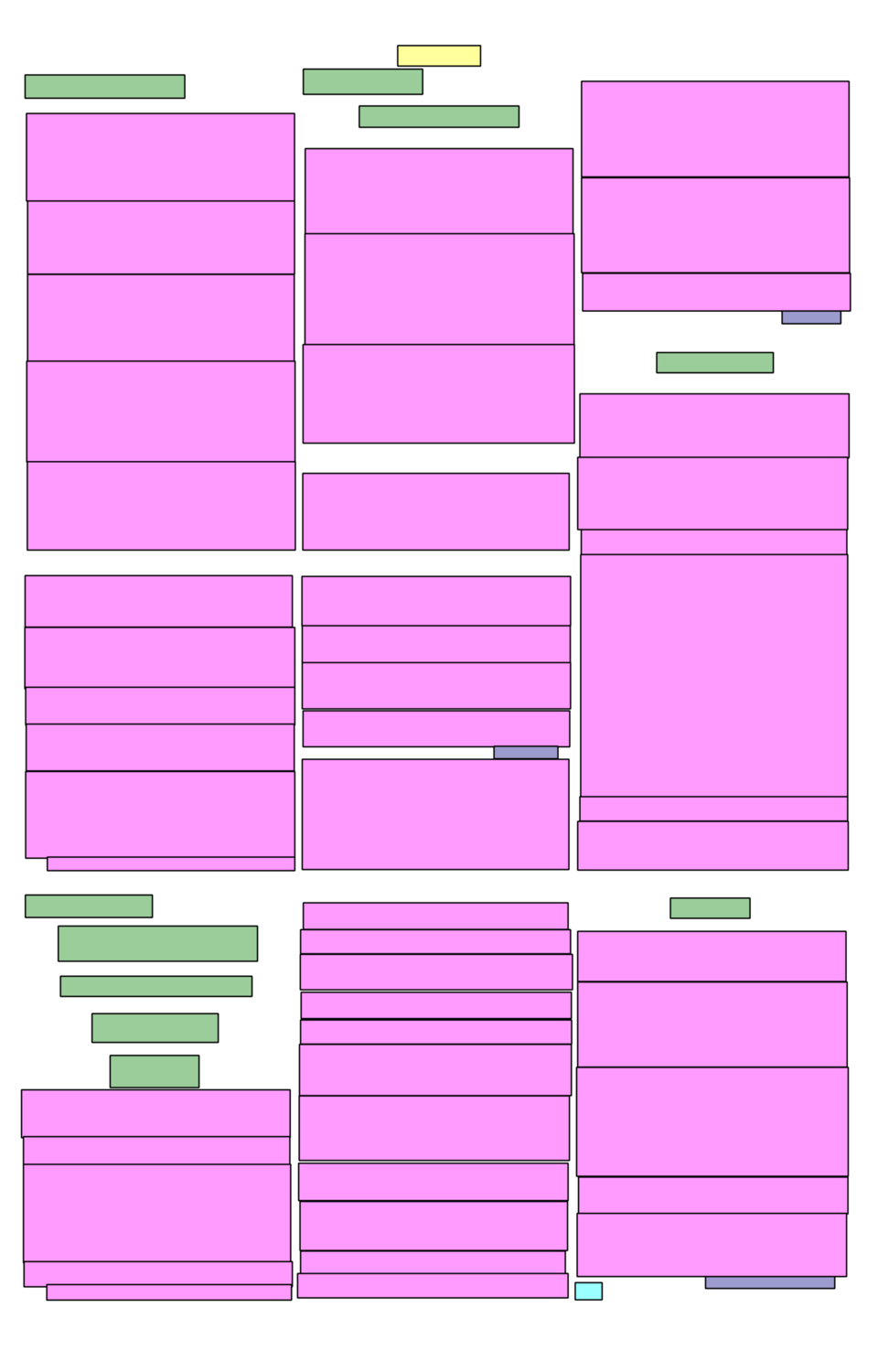} &
        \includegraphics[width=0.22\textwidth]{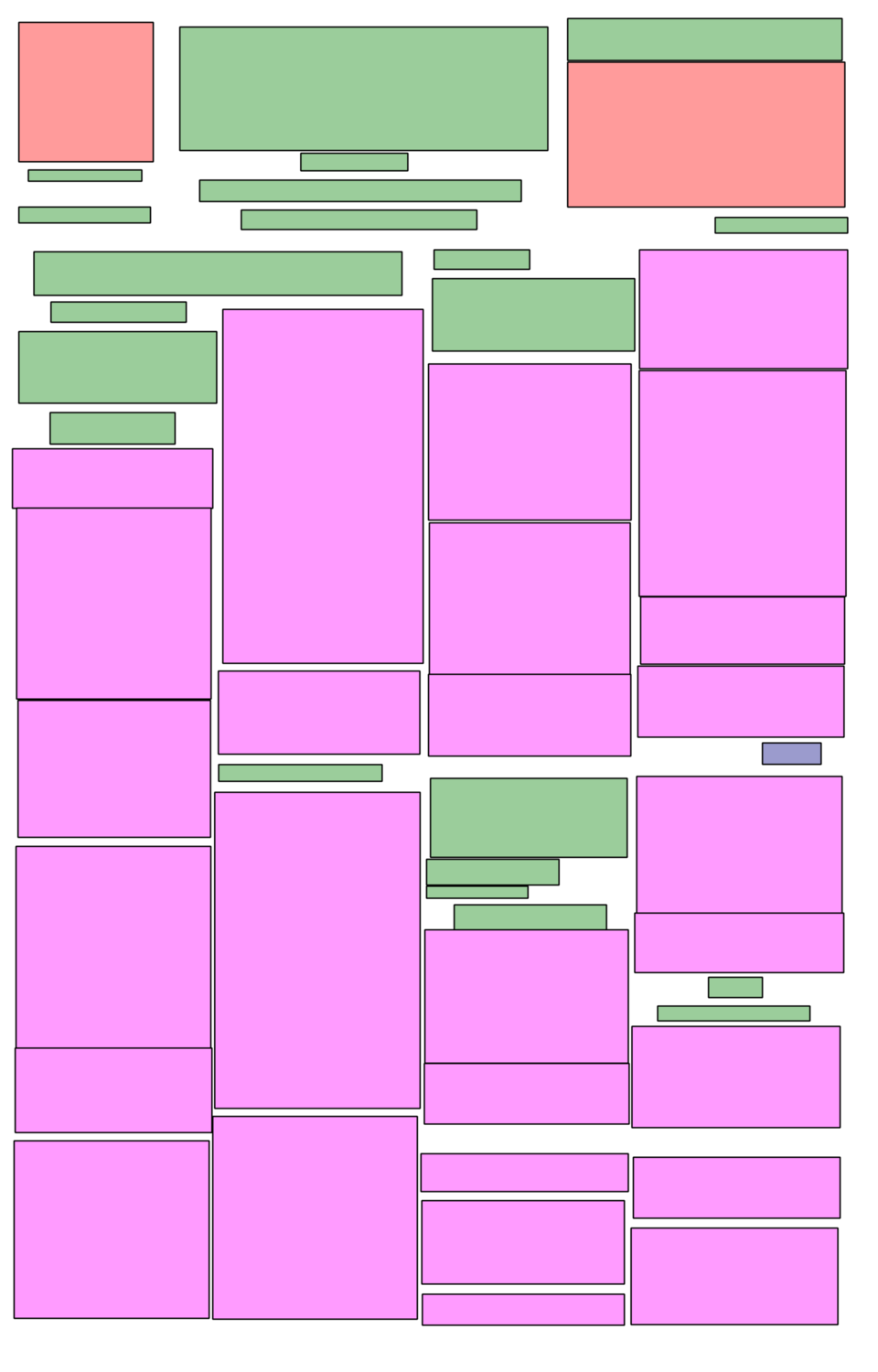} &
        \includegraphics[width=0.22\textwidth]{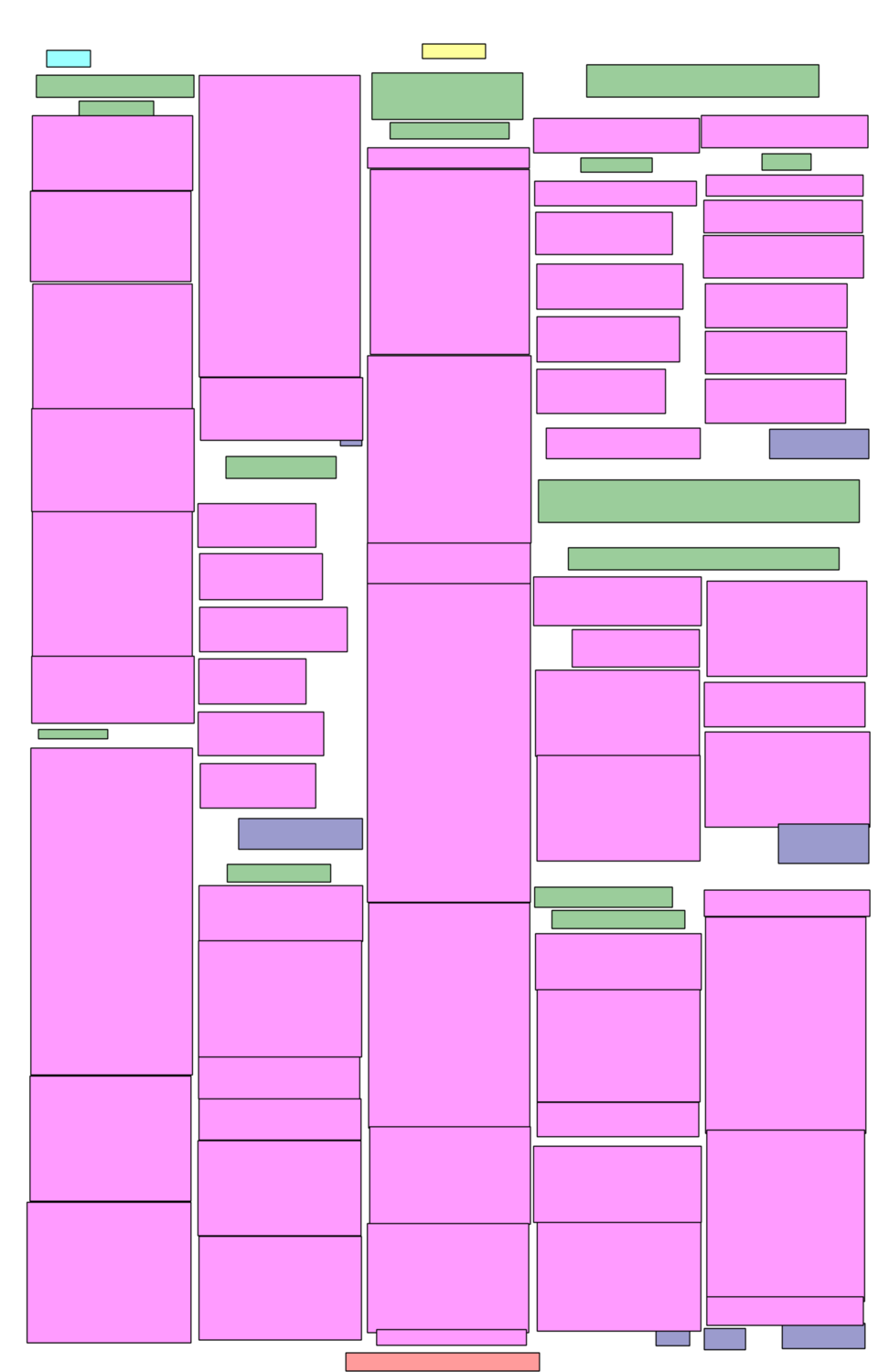} \\

        \textbf{Semantic blocks} &
        \includegraphics[width=0.22\textwidth]{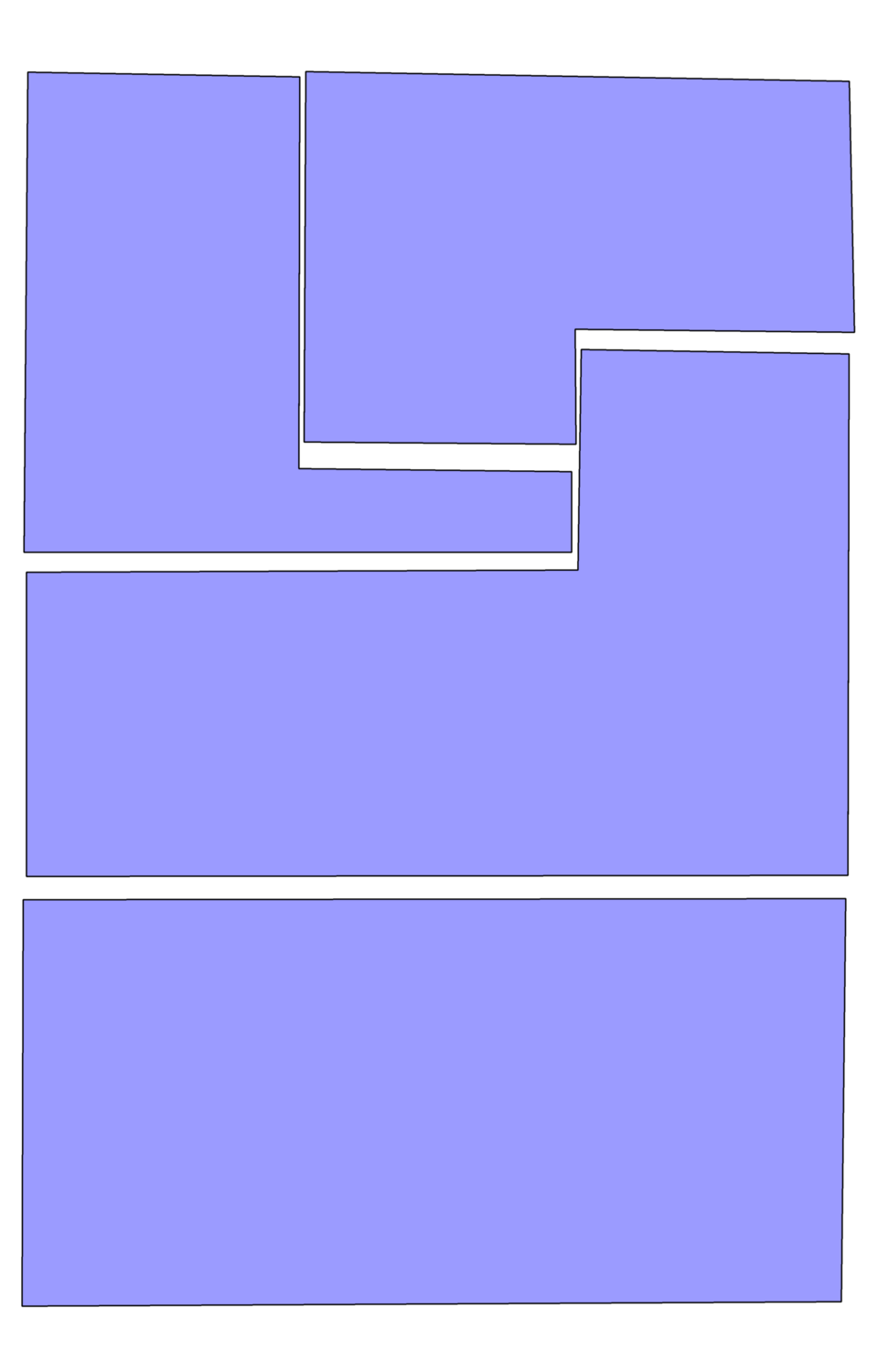} &
        \includegraphics[width=0.22\textwidth]{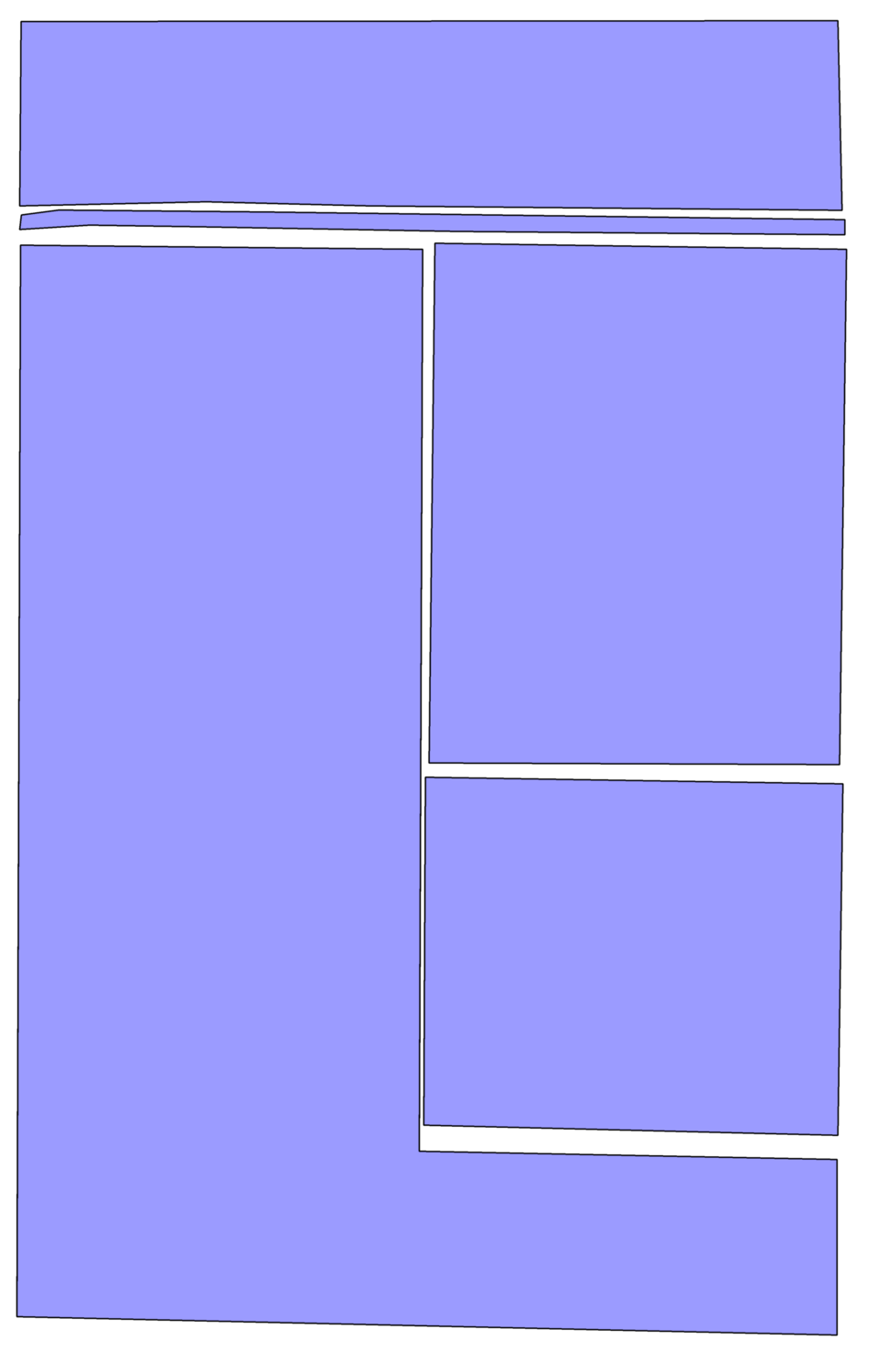} &
        \includegraphics[width=0.22\textwidth]{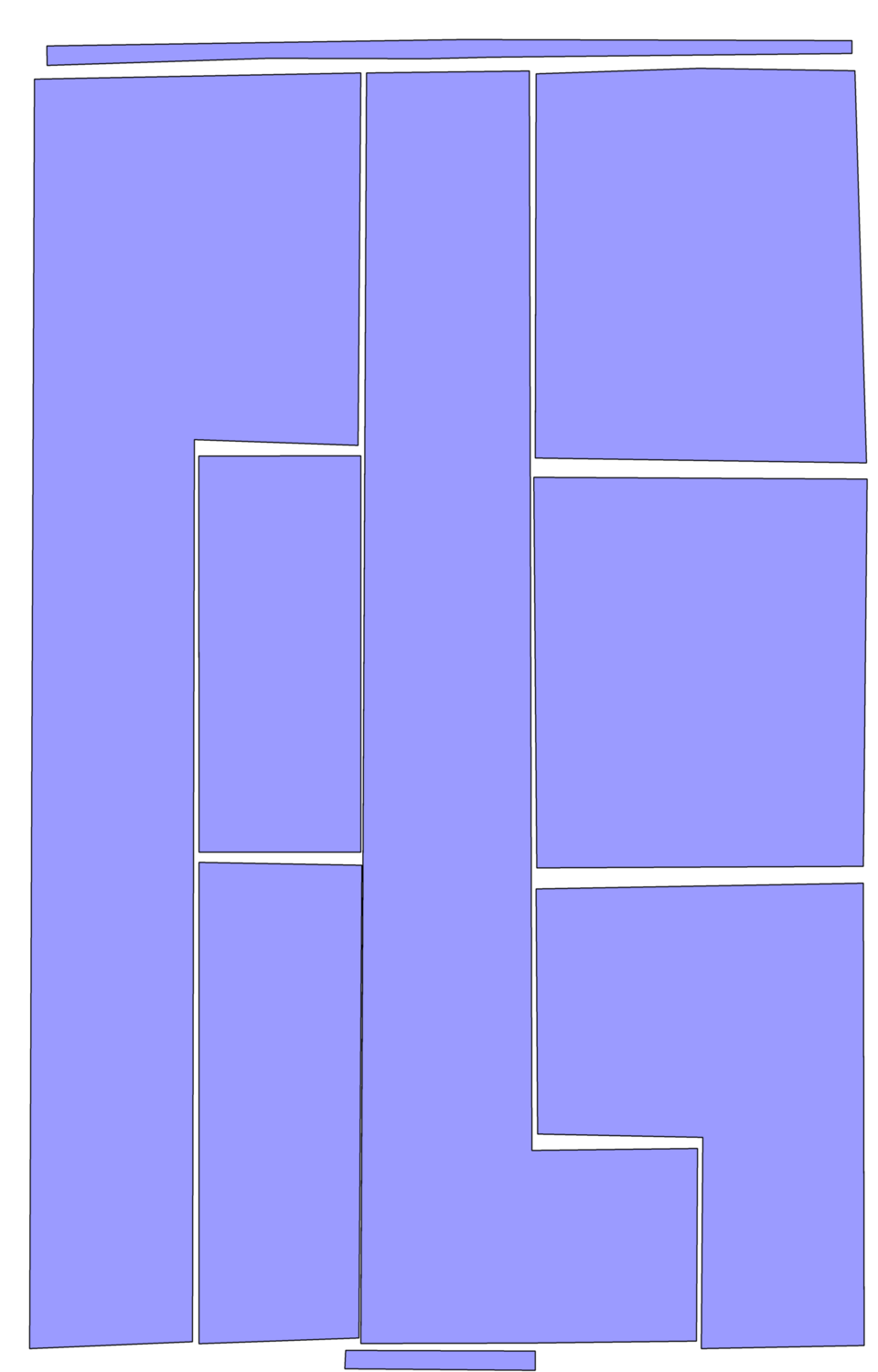} \\

        \textbf{Separators} &
        \includegraphics[width=0.22\textwidth]{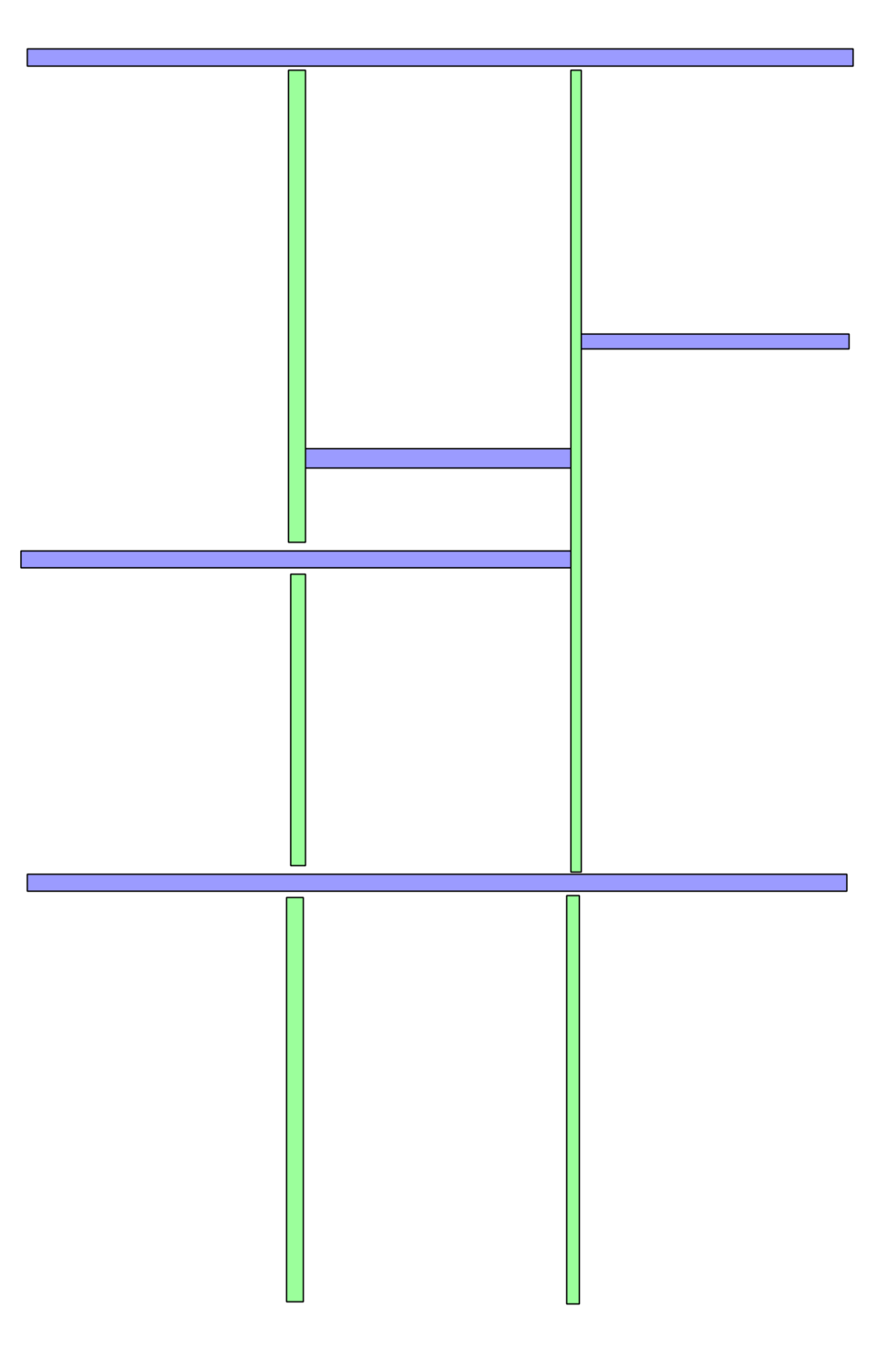} &
        \includegraphics[width=0.22\textwidth]{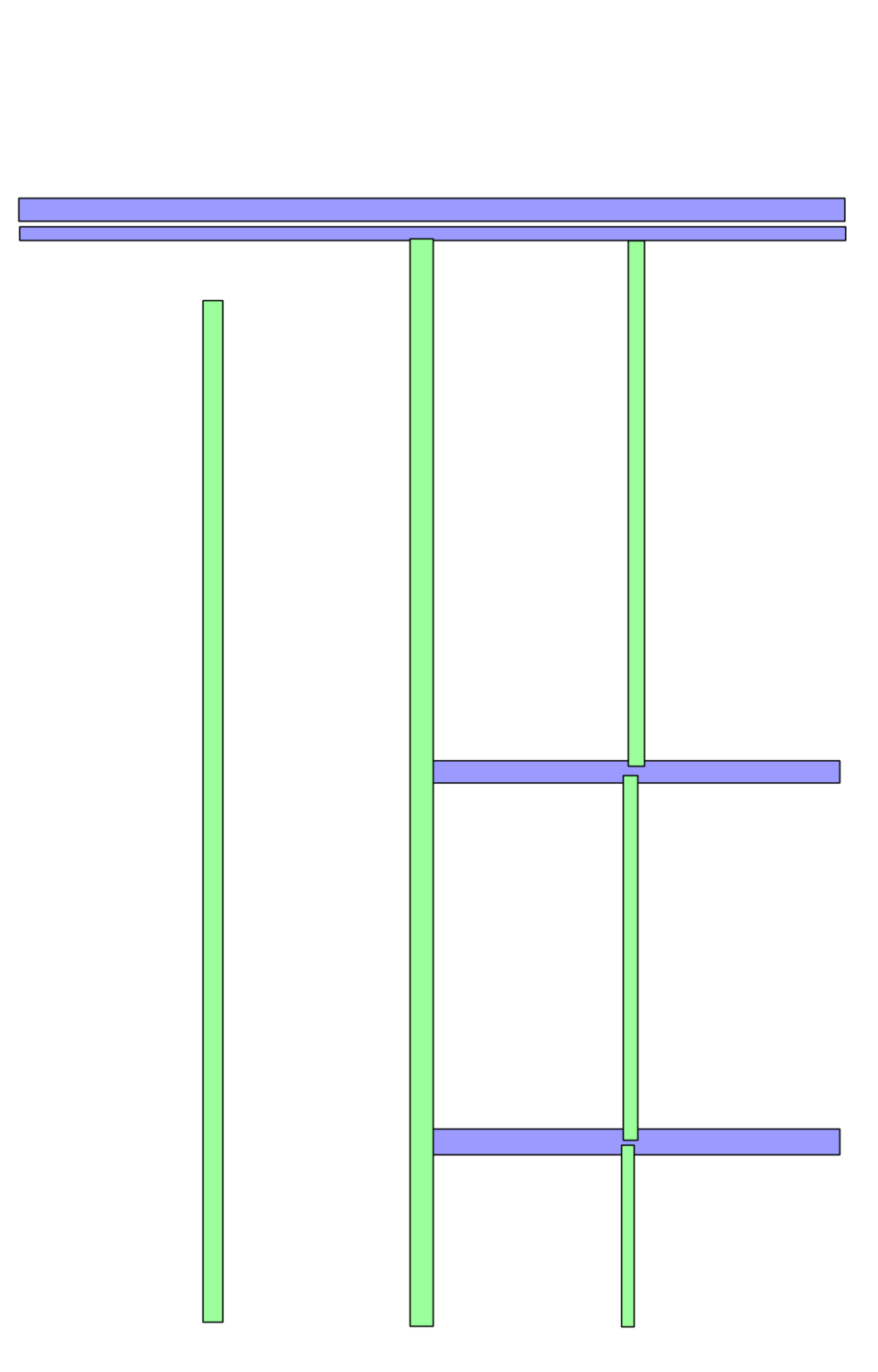} &
        \includegraphics[width=0.22\textwidth]{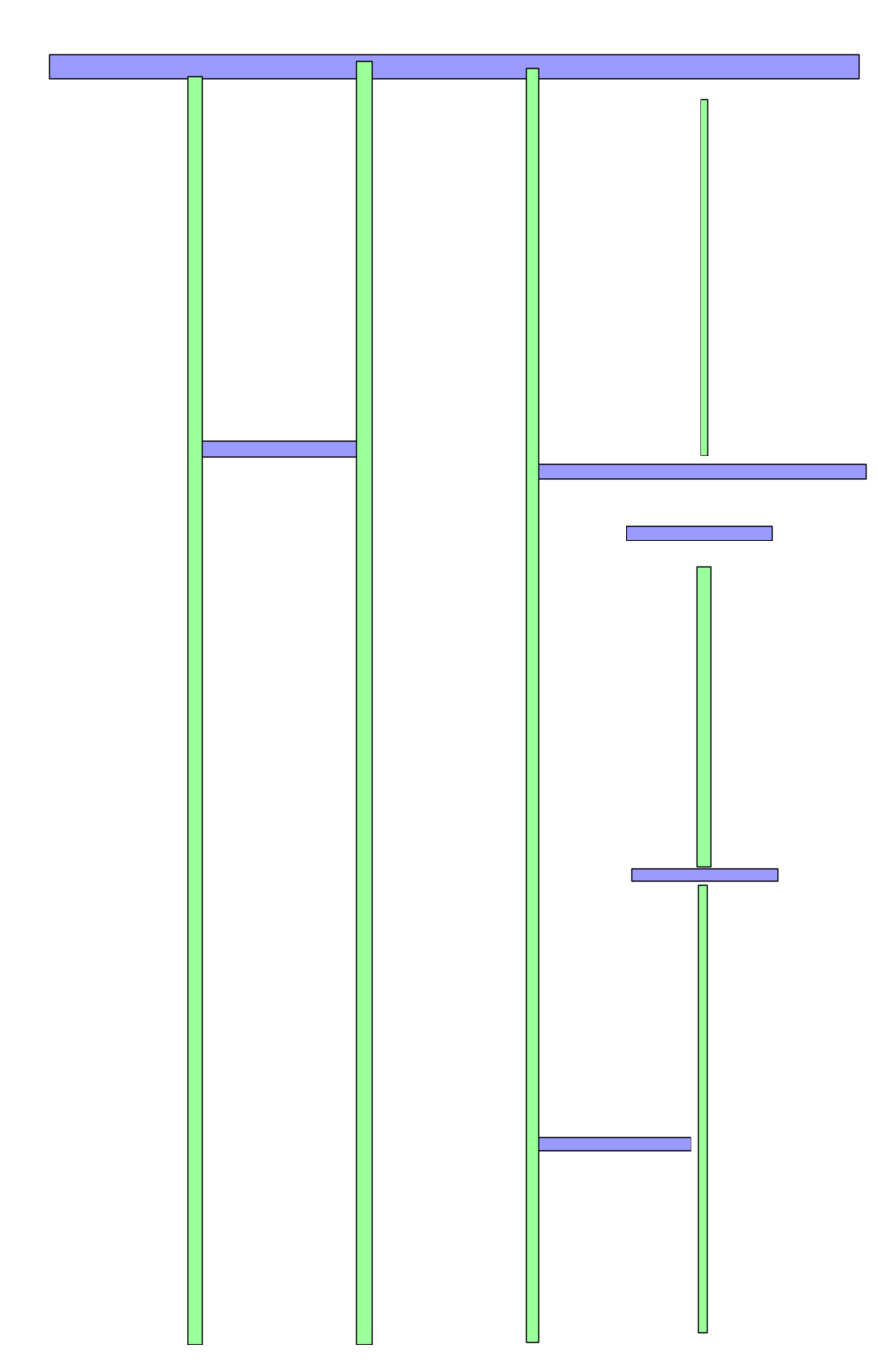} \\
    \end{tabular}%
    }
    \caption{Examples of original images, text region annotations, semantic blocks, and structural separators. Source: BULAC and ARAM.}
    \label{tab:visual_examples}
\end{table}

\section{Tasks and experiments}\label{sec:tasks}
We use the dataset for two reading order tasks. \textbf{Task 1} addresses reading order within a single page; \textbf{Task 2} identifies the global reading sequence across multiple pages, a common newspaper scenario where an article begins on the front page and continues on later ones. \new{Fig.~\ref{fig:tasks} illustrates both.}

\subsection{Preliminary task about OCR}

\begin{figure}[!htbp]
\centering
\resizebox{\textwidth}{!}{%
\begin{tikzpicture}[
    node distance=3mm,
    every node/.style={draw, rounded corners, align=center, minimum width=10mm, minimum height=8mm, font=\scriptsize},
    conv/.style={fill=blue!20},
    pool/.style={fill=cyan!20},
    lstm/.style={fill=orange!25},
    ctc/.style={fill=red!25},
    input/.style={fill=gray!15},
    arrow/.style={->, line width=0.5pt}
]

\node[input] (input) {Input \\ (1 channel)};
\node[conv, right=of input] (conv)  {Conv 5×5 \\ 16 filters};
\node[pool, right=of conv] (pool)  {Pooling 3×3};
\node[lstm, right=of pool] (lstmY) {BiLSTM$_{y}$ \\ 64 units};
\node[lstm, right=of lstmY] (lstmXf) {LSTM$_{x}$ \\ 128};
\node[lstm, right=of lstmXf] (lstmXr) {LSTM$_{x}$ (rev) \\ 128};
\node[lstm, right=of lstmXr] (lstm) {LSTM$_{x}$ \\ 256};
\node[ctc, right=of lstm] (ctc) {CTC};

\draw[arrow] (input) -- (conv);
\draw[arrow] (conv) -- (pool);
\draw[arrow] (pool) -- (lstmY);
\draw[arrow] (lstmY) -- (lstmXf);
\draw[arrow] (lstmXf) -- (lstmXr);
\draw[arrow] (lstmXr) -- (lstm);
\draw[arrow] (lstm) -- (ctc);

\end{tikzpicture}%
}
\caption{VGSL OCR architecture with a CTC decoder. This compact model is well-suited to our $\sim$15{,}000-line training set; the stacked LSTM layers implicitly capture local sequential dependencies (a lightweight LM substitute).}
\label{fig:vgsl_schema}
\end{figure}
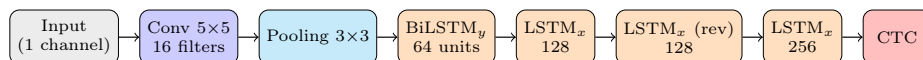

Using LLMs requires transcriptions, and Armenian OCR is already difficult on its own: open models Tesseract and EasyOCR perform poorly on historical fonts and low-resolution scans, and their internal language models cover Western Armenian sparsely. A generic Armenian OCR/HTR model has been proposed by Calfa~\cite{vidal2025armenian,chague:hal-05163931}, but it targets handwritten archives and ancient manuscripts. We therefore train our own Tesseract-based model on the dataset of this paper (CRNN, $<$300k parameters; see Fig.~\ref{fig:vgsl_schema}; average inference 1.3s per page, $\sim$0.1s per line), tailored to rare typefaces and variable scan quality.

Absolute CER reductions range from 8 to 35 percentage points (Table~\ref{tab:ocr_results}), with the largest gains on noisy and damaged historical fonts; most residual errors localize on non-Armenian words and curved lines.

\subsection{Preliminary task about layout analysis}

YOLO-family models perform strongly on complex and historical layouts~\cite{sven2022page}. We compare YOLOv11x (default hyperparameters, dropout 0.1, image size 1,560) and DocLayout-YOLO~\cite{zhao2024doclayout}, a layout-aware model pretrained on the synthetic DocSynth dataset. Table~\ref{tab:layout_analysis} shows that DocLayout-YOLO + SynthData consistently achieves the best results, confirming the benefit of combining a layout-specific architecture with synthetic pretraining; DocLayout-YOLO outperforms YOLOv11x on mAP@50 and recall by a small margin (1--2 pp), while YOLOv11x remains competitive when fine-tuned.

\begin{table}[!htbp]
\centering
\small
\renewcommand{\arraystretch}{1.1}
\begin{tabularx}{\textwidth}{p{5.3cm} >{\raggedleft\arraybackslash}X >{\raggedleft\arraybackslash}X >{\raggedleft\arraybackslash}X}
\toprule
\textbf{Example} & \textbf{Metric} & \textbf{Default} & \textbf{Specialized} \\
\midrule
\raisebox{-0.5\totalheight}{\includegraphics[width=1.5\linewidth]{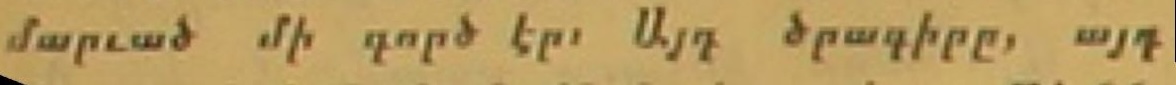}} \\ 
\textit{Blurry 20th c. newspaper} (-20\% err.) &
CER & 28.95 & 8.61 \\
& WER & 95.96 & 52.22 \\
\midrule
\raisebox{-0.5\totalheight}{\includegraphics[width=1.5\linewidth]{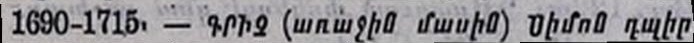}} \\
\textit{Noisy 20th c. newspaper} (-28\%) &
CER & 36.64 & 8.11 \\
& WER & 101.22 & 44.38 \\
\midrule
\raisebox{-0.5\totalheight}{\includegraphics[width=1.5\linewidth]{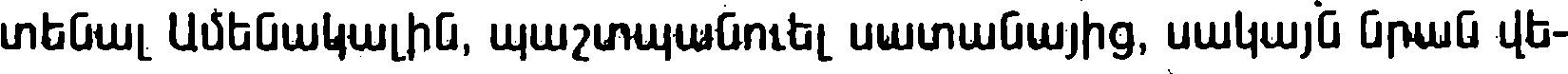}} \\
\textit{Binarized newspaper} (-8\%) &
CER & 11.75 & 3.99 \\
& WER & 50.07 & 21.51 \\
\midrule
\raisebox{-0.5\totalheight}{\includegraphics[width=1.5\linewidth]{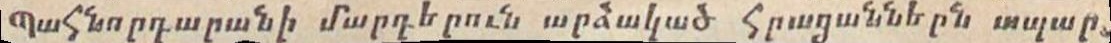}} \\
\textit{Historical damaged font} (-35\%) &
CER & 48.93 & 13.47 \\
& WER & 134.90 & 67.63 \\
\bottomrule
\end{tabularx}
\caption{Out-of-domain CER/WER of default vs.\ specialized Tesseract on four challenging Armenian line types.}
\label{tab:ocr_results}
\end{table}

Synthetic data improves both models, especially for harder classes (\texttt{Marginalia}, \texttt{MetaData}); gains are more pronounced for YOLOv11x, which benefits more from layout exposure than the already-pretrained DocLayout-YOLO. \texttt{Marginalia}, \texttt{MetaData}, \texttt{Signature}, and \texttt{Advertisement} remain poorly detected due to their rarity and variability—acceptable here as they are peripheral to reading order. The classes that matter most for our task, \texttt{Title} and \texttt{Paragraph}, are reliably detected across all models.

\begin{table}[!htbp]
\centering
\small
\renewcommand{\arraystretch}{1.2}
\resizebox{\textwidth}{!}{%
\begin{tabular}{llcccccccccc}
\textbf{Model} & & 
\rotatebox{90}{All} & 
\rotatebox{90}{Advertisement} & 
\rotatebox{90}{Marginalia} & 
\rotatebox{90}{MetaData} & 
\rotatebox{90}{PageNum.} & 
\rotatebox{90}{Paragraph} & 
\rotatebox{90}{Run.Title} & 
\rotatebox{90}{Sign.} & 
\rotatebox{90}{Title} \\
\midrule
\multirow{3}{*}{YOLOv11x} 
& P     & 0.720 & 0.758 & 0.383 & 0.574 & 0.772 & 0.958 & 0.741 & 0.804 & 0.767 \\
& R     & 0.686 & \textbf{1.000} & 0.100 & 0.539 & 0.521 & 0.912 & 0.880 & 0.760 & 0.780 \\
& mAP50 & 0.739 & 0.995 & 0.202 & 0.603 & 0.627 & 0.971 & 0.871 & 0.811 & 0.830 \\
\midrule
\multirow{3}{*}{YOLOv11x + Synth} 
& P     & 0.789 & 0.854 & 0.437 & 0.652 & 0.807 & 0.988 & 0.837 & 0.893 & 0.842 \\
& R     & 0.720 & \textbf{1.000} & 0.109 & 0.546 & 0.558 & 0.958 & 0.931 & 0.823 & 0.832 \\
& mAP50 & 0.798 & \textbf{1.000} & 0.203 & 0.648 & 0.706 & \textbf{0.981} & \textbf{0.946} & 0.865 & 0.908 \\
\midrule
\multirow{3}{*}{DocLayout-YOLO} 
& P     & 0.779 & 0.807 & 0.406 & 0.606 & 0.829 & \textbf{0.989} & 0.810 & 0.877 & 0.829 \\
& R     & 0.728 & \textbf{1.000} & 0.110 & 0.581 & 0.550 & \textbf{0.970} & 0.929 & 0.826 & 0.829 \\
& mAP50 & 0.787 & \textbf{1.000} & \textbf{0.215} & 0.651 & 0.701 & \textbf{0.981} & 0.910 & 0.861 & 0.866 \\
\midrule
\multirow{3}{*}{DocLayout-YOLO + Synth} 
& P     & \textbf{0.831} & \textbf{0.882} & \textbf{0.444} & \textbf{0.662} & \textbf{0.881} & 0.971 & \textbf{0.871} & \textbf{0.954} & \textbf{0.902} \\
& R     & \textbf{0.779} & \textbf{1.000} & \textbf{0.118} & \textbf{0.608} & \textbf{0.598} & 0.912 & \textbf{1.000} & \textbf{0.891} & \textbf{0.872} \\
& mAP50 & \textbf{0.847} & \textbf{1.000} & \textbf{0.232} & \textbf{0.693} & \textbf{0.751} & 0.961 & \textbf{0.983} & \textbf{0.936} & \textbf{0.931} \\
\bottomrule
\end{tabular}
} 
\caption{Out-of-domain semantic text region detection results.}
\label{tab:layout_analysis}
\end{table}

\begin{algorithm}[H]
\DontPrintSemicolon
\KwIn{List of bounding boxes $\mathcal{B} = \{(x_1, y_1, x_2, y_2)\}$, image shape $(H, W)$, band-height threshold $T$ (median paragraph height)}
\KwOut{Assigned \texttt{paragraph\_id} and \texttt{row\_id} for each box}

Initialize a blank binary heatmap of shape $(H, W)$\;
\ForEach{$b_i \in \mathcal{B}$}{
    Fill corresponding region in heatmap with $0$
}

Compute horizontal projection $\mathcal{H}(y)$ of heatmap\;
Find zero spans in $\mathcal{H}(y)$ to detect horizontal breakpoints\;
Split image into horizontal bands using these breakpoints\;

\ForEach{band}{
    \eIf{band height $< T$}{
        Assign all boxes to one paragraph group\;
    }{
        Compute vertical projection $\mathcal{V}(x)$ within band\;
        Find vertical breakpoints via peak detection in $-\mathcal{V}(x)$\;
        Divide band into column zones using these breakpoints\;
        Assign paragraph groups by region\;
    }
}

Merge overlapping paragraphs based on spatial IoU\;
Assign \texttt{row\_id}s within each paragraph using vertical overlap\;
Refine grouping by detecting overlong rows and splitting heuristically\;
\Return{updated bounding boxes with \texttt{paragraph\_id} and \texttt{row\_id}}
\caption{Whitespace-based layout grouping with YOLO bounding boxes}
\label{alg:yolo_layout_refined}
\end{algorithm}

\subsection{Reading order strategies}

\paragraph{Global Topological Sort (GTS).}
Inspired by Breuel~\cite{breuel2003high}, we adapt a line-segment topological method to paragraph-level bounding boxes: layout elements are segmented from geometric criteria (baseline alignment, spacing, font size) and the reading order is recovered as a topological sort over the resulting partial order, $\mathcal{O} = \text{TopSort}(\{b_i\},\{b_i \prec b_j\})$, robust to multi-column layouts and floating elements.

\paragraph{Whitespace deduction + LTS (W~ded.\,+\,LTS).}
For boxes $\mathcal{B}=\{b_i\}$ predicted by YOLO, we compute horizontal and vertical projections
\[
\mathcal{H}(y) = \sum_{i} \mathbf{1}_{[y_{1,i}, y_{2,i}]}(y), \quad
\mathcal{V}(x) = \sum_{i} \mathbf{1}_{[x_{1,i}, x_{2,i}]}(x),
\]
and \emph{deduce} whitespace separators as long zero-valued spans after smoothing (Alg.~\ref{alg:yolo_layout_refined}, with band-height threshold $T$ set to the median paragraph height). The procedure assigns each box a \texttt{paragraph\_id} and a hierarchical \texttt{row\_id}; reading order is then produced by a Local Topological Sort (LTS) within each group, ordering by \texttt{row\_id} and then by horizontal position. Adapted from whitespace-based segmentation~\cite{smith2009hybrid,breuel2003layout} but applied to detector outputs, this is robust to prediction noise.

\paragraph{Whitespace detection + LTS (W~det.\,+\,LTS).}
As an alternative to deducing separators from box projections, we train a dedicated YOLO model to \emph{detect} horizontal and vertical separators directly (more reliable when boxes do not align cleanly with paragraph structure), then run the same grouping and LTS pipeline.

\paragraph{MLP.}
Adapting~\cite{quiros2018multi}, we train a Multi-Layer Perceptron on YOLO-detected paragraph boxes with ground-truth annotations to predict reading order from spatial features—a learned alternative to topological sorting.

\paragraph{SD + LTS.}
The two-pass YOLO above (semantic blocks then paragraphs) is also evaluated without an LLM: paragraphs are grouped by semantic block and ordered within each block by a Local Topological Sort, then blocks are concatenated top-to-bottom. This isolates the contribution of semantic grouping from that of the LLM.

Inspired by hierarchical detectors for complex Chinese reading order~\cite{bizais2024optimizing}, we also use a two-pass YOLO that first detects semantic blocks (Table~\ref{tab:visual_examples}) and then individual paragraphs—the basis of our SD\,+\,LLM pipeline below.

\begin{figure}[!htbp]
\small
\centering
\begin{promptbox}[Reading Order Inference (Armenian newspaper)]
\textbf{Role:} You are an expert in Armenian language.

\textbf{Task:} Reorder the following paragraphs from a single Armenian newspaper page that are out of order (raw OCR output).
\begin{itemize}
\item Use pairwise contextual coherence to determine the most logical sequence.  
\item Preserve every word exactly; only change the paragraph order.
\item Don't reorder sentences inside a paragraph.
\item Return the reordered paragraphs, each on its own line, with no extra commentary.
\item Input is a raw line-level OCR output and can contain typos, make sure to take into account the overall context. Hyphens can be a relevant clue to help.
\end{itemize}

\textbf{Example 1 (need re-ordering):}\\
Unordered input: [Block1] [Block3] [Block2] [Block4]

Expected output: [Block1] [Block2] [Block3] [Block4]

\textbf{Example 2 (doesn't need re-ordering):}\\
Unordered input: [Block1] [Block2] [Block3] [Block4]

Expected output (no change): [Block1] [Block2] [Block3] [Block4]
\end{promptbox}
\caption{Example of LLM prompt used to infer paragraph reading order from OCR output.}
\label{fig:prompt}
\end{figure}

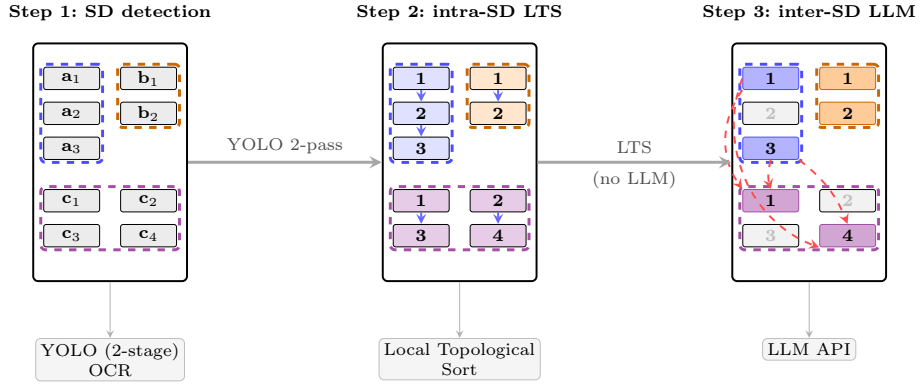
\begin{figure}[!htbp]
\centering
{
\resizebox{\textwidth}{!}{%
\begin{tikzpicture}[
  font=\small,
  page/.style={draw, thick, rounded corners=2pt, minimum width=22mm, minimum height=34mm},
  para/.style={draw, fill=gray!15, rounded corners=1pt, font=\scriptsize, minimum width=8mm, minimum height=3.2mm, inner sep=1pt},
  sda/.style={draw=blue!70, very thick, dashed, rounded corners=2pt, inner sep=1pt},
  sdb/.style={draw=orange!80!black, very thick, dashed, rounded corners=2pt, inner sep=1pt},
  sdc/.style={draw=violet!70, very thick, dashed, rounded corners=2pt, inner sep=1pt},
  arrgeo/.style={->, >=stealth, thick, blue!60},
  arrllm/.style={->, >=stealth, thick, red!70, dashed},
  stepl/.style={font=\scriptsize\bfseries}
]

\node[page] (P1) at (0,0) {};
\node[above=2mm of P1, stepl] {Step 1: SD detection};

\node[para] (p1a) at (-0.55, 1.20) {\textbf{a$_1$}};
\node[para] (p1b) at (-0.55, 0.70) {\textbf{a$_2$}};
\node[para] (p1c) at (-0.55, 0.20) {\textbf{a$_3$}};
\node[sda, fit=(p1a)(p1c)] {};

\node[para] (p2a) at ( 0.55, 1.20) {\textbf{b$_1$}};
\node[para] (p2b) at ( 0.55, 0.70) {\textbf{b$_2$}};
\node[sdb, fit=(p2a)(p2b)] {};

\node[para] (p3a) at (-0.55, -0.55) {\textbf{c$_1$}};
\node[para] (p3b) at ( 0.55, -0.55) {\textbf{c$_2$}};
\node[para] (p3c) at (-0.55, -1.05) {\textbf{c$_3$}};
\node[para] (p3d) at ( 0.55, -1.05) {\textbf{c$_4$}};
\node[sdc, fit=(p3a)(p3d)] {};

\node[page] (P2) at (5,0) {};
\node[above=2mm of P2, stepl] {Step 2: intra-SD LTS};

\node[para, fill=blue!12]   (q1a) at (4.45, 1.20) {\textbf{1}};
\node[para, fill=blue!12]   (q1b) at (4.45, 0.70) {\textbf{2}};
\node[para, fill=blue!12]   (q1c) at (4.45, 0.20) {\textbf{3}};
\draw[arrgeo] (q1a) -- (q1b); \draw[arrgeo] (q1b) -- (q1c);

\node[para, fill=orange!20] (q2a) at (5.55, 1.20) {\textbf{1}};
\node[para, fill=orange!20] (q2b) at (5.55, 0.70) {\textbf{2}};
\draw[arrgeo] (q2a) -- (q2b);

\node[para, fill=violet!20] (q3a) at (4.45, -0.55) {\textbf{1}};
\node[para, fill=violet!20] (q3b) at (5.55, -0.55) {\textbf{2}};
\node[para, fill=violet!20] (q3c) at (4.45, -1.05) {\textbf{3}};
\node[para, fill=violet!20] (q3d) at (5.55, -1.05) {\textbf{4}};
\draw[arrgeo] (q3a) -- (q3c); \draw[arrgeo] (q3b) -- (q3d);

\node[sda, fit=(q1a)(q1c)] {};
\node[sdb, fit=(q2a)(q2b)] {};
\node[sdc, fit=(q3a)(q3d)] {};

\node[page] (P3) at (10,0) {};
\node[above=2mm of P3, stepl] {Step 3: inter-SD LLM};

\node[para, fill=blue!30, draw=blue!70] (r1a) at (9.45, 1.20) {\textbf{1}};
\node[para, fill=gray!10, text=gray!50] (r1b) at (9.45, 0.70) {\textbf{2}};
\node[para, fill=blue!30, draw=blue!70] (r1c) at (9.45, 0.20) {\textbf{3}};
\node[sda, fit=(r1a)(r1c)] {};

\node[para, fill=orange!40, draw=orange!80!black] (r2a) at (10.55, 1.20) {\textbf{1}};
\node[para, fill=orange!40, draw=orange!80!black] (r2b) at (10.55, 0.70) {\textbf{2}};
\node[sdb, fit=(r2a)(r2b)] {};

\node[para, fill=violet!35, draw=violet!70] (r3a) at (9.45, -0.55) {\textbf{1}};
\node[para, fill=gray!10, text=gray!50]      (r3b) at (10.55, -0.55) {\textbf{2}};
\node[para, fill=gray!10, text=gray!50]      (r3c) at (9.45, -1.05) {\textbf{3}};
\node[para, fill=violet!35, draw=violet!70]  (r3d) at (10.55, -1.05) {\textbf{4}};
\node[sdc, fit=(r3a)(r3d)] {};

\draw[arrllm] (r1a.south west) to[bend right=35] (r3a.north west);
\draw[arrllm] (r1a.west)       to[bend right=45] (r3d.south west);
\draw[arrllm] (r1c.south)      to[bend right=15] (r3a.north);
\draw[arrllm] (r1c.south east) to[bend left=15]  (r3d.north);

\draw[->, >=stealth, very thick, gray!70] (P1.east) -- node[above, font=\scriptsize, text=gray!60!black] {YOLO 2-pass} (P2.west);
\draw[->, >=stealth, very thick, gray!70] (P2.east) -- node[above, font=\scriptsize, text=gray!60!black] {LTS} node[below, font=\scriptsize, text=gray!60!black] {(no LLM)} (P3.west);

\node[draw=gray!50, fill=gray!8, rounded corners=2pt, font=\scriptsize, align=center, inner sep=2pt, below=8mm of P1] (mod1) {YOLO (2-stage)\\OCR};
\node[draw=gray!50, fill=gray!8, rounded corners=2pt, font=\scriptsize, align=center, inner sep=2pt, below=8mm of P2] (mod2) {Local Topological\\Sort};
\node[draw=gray!50, fill=gray!8, rounded corners=2pt, font=\scriptsize, align=center, inner sep=2pt, below=8mm of P3] (mod3) {LLM API};

\draw[->, >=stealth, gray!60] (P1.south) -- (mod1.north);
\draw[->, >=stealth, gray!60] (P2.south) -- (mod2.north);
\draw[->, >=stealth, gray!60] (P3.south) -- (mod3.north);

\end{tikzpicture}%
}
}
\caption{\new{Proposed SD\,+\,Generative LLM pipeline: SD detection, intra-SD Local Topological Sort, then LLM pairwise comparison restricted to the first and last paragraphs of each SD. Dashed red arrows are shown only between SD$_1$--SD$_3$ for readability; all SD pairs are compared.}}
\label{fig:pipeline}
\end{figure}

We benchmark ECLAIR~\cite{karmanov2025eclair} alongside our proposed SD\,+\,Generative LLM pipeline (Fig.~\ref{fig:pipeline}). The LLM (Mistral-8B-Instruct-2410) is queried under a role-play system prompt with two-shot in-context learning (Fig.~\ref{fig:prompt}): one example requires reordering, the other is already in correct order, so the model learns when \emph{not} to act. The pipeline runs in three steps: (i) two-pass YOLO yields SDs and their paragraphs; (ii) paragraphs within each SD are ordered by a Local Topological Sort (LTS, no LLM call); (iii) the global order across SDs is decided by the LLM through pairwise comparisons restricted to the first and last paragraphs of each pair of SDs—reducing LLM calls from $O(n^2)$ to $O(k^2)$ with $k \ll n$. All runs use temperature $0$ ($\sim$8\,s/page on one GPU). We additionally evaluate Gemini~3.0~Flash as a drop-in replacement for Mistral-8B to test the sensitivity of the pipeline to the underlying LLM.

\section{Results and discussions}

\subsection{Metrics}

Following Quirós \& Vidal~\cite{quiros2022reading}, we use Kendall's tau distance $\tau$ (number of pairwise inversions) and Spearman's footrule $F$ (sum of absolute positional differences), defined as
\begin{equation}
\tau = \frac{2K}{n(n-1)}, \qquad F = \frac{\sum_{i=1}^{n} |\pi(i) - \sigma(i)|}{\left\lfloor n^2/2 \right\rfloor},
\end{equation}
where $K$ is the number of inverted pairs, $n$ the number of elements, and $\pi(i), \sigma(i)$ the predicted and ground-truth ranks of item $i$. Both are normalized to $[0,1]$ with $0 = $ perfect agreement.

{\paragraph*{Metric scope and independent articles.}
Both metrics assume a single fully-ordered ground truth, but newspaper pages routinely carry several independent articles whose relative order is arbitrary: reading column~2 before column~1 is equally valid. Such inter-article permutations are penalized identically to genuine intra-article errors. Scores should therefore be read as an upper bound on true ordering errors; the limitation is inherent to single-sequence evaluation and affects all methods equally.}

\noindent\textbf{Abbreviations (Tables~\ref{tab:task1_results} and~\ref{tab:llm_comparison}).}
\textbf{GTS}: Global Topological Sort.
\textbf{W ded. \, + \, LTS}: Whitespace deduction + Local Topological Sort.
\textbf{W det. \, + \, LTS}: YOLO separator detection + LTS.
\textbf{MLP}: Multi-Layer Perceptron on bounding-box features.
\textbf{SD\,+\,LTS}: Semantic block Detection + LTS.
\textbf{ECLAIR}: end-to-end OCR/layout model (ViT-H + mBART)~\cite{karmanov2025eclair}.
\textbf{Gen. LLM}: Mistral-8B-Instruct-2410, prompt-only.
\textbf{SD\,+\,Gen. LLM}: SD + Generative LLM \textit{(proposed)}.
\textbf{SD\,+\,Gemini}: SD + Gemini~3.0~Flash \textit{(ablation)}.

\begin{table}[!htbp]
\centering
\small
\caption{Reading Order Accuracy – Task 1}
\resizebox{\textwidth}{!}{%
\begin{tabular}{lcccc}
\toprule
\textbf{Method} & \multicolumn{2}{c}{\textbf{Exp. 1: Shuffled GT}} & \multicolumn{2}{c}{\textbf{Exp. 2: Predicted BBoxes}} \\
\cmidrule(lr){2-3} \cmidrule(lr){4-5}
 & $\tau$ ↓ & $F$ ↓ & $\tau$ ↓ & $F$ ↓ \\
\midrule
GTS & 0.35 & 0.55 & 0.42 & 0.50 \\
W ded. + LTS & 0.25 & 0.45 & 0.38 & 0.43 \\
W det. + LTS & 0.22 & 0.42 & 0.33 & 0.39 \\
MLP & 0.70 & 0.68 & 0.73 & 0.66 \\
SD + LTS & 0.14 & 0.38 & 0.17 & 0.35 \\
ECLAIR & 0.18 & 0.40 & 0.22 & 0.38 \\
Generative LLM (Mistral-8B) & 0.75 & 0.82 & 0.79 & 0.85 \\
SD + Generative LLM (Mistral-8B) & 0.07 & 0.20 & 0.10 & 0.18 \\
SD + Generative LLM (other variants) & \multicolumn{4}{c}{\textit{see Table~\ref{tab:llm_comparison}}} \\
\bottomrule
\end{tabular}%
}
\label{tab:task1_results}
\end{table}

\subsection{Reading Order Evaluation Results}

Table~\ref{tab:task1_results} reports $\tau$ and $F$ (both in $[0,1]$, lower is better) under two settings: Exp.~1 randomly shuffles ground-truth bounding boxes; Exp.~2 uses predicted boxes from detection and recognition. \new{LLM variants within the SD\,+\,Generative LLM pipeline are compared separately in Table~\ref{tab:llm_comparison}.}

SD + Generative LLM performs best in both experiments, achieving the lowest scores (0.07/0.20 in Exp.~1 and 0.10/0.18 in Exp.~2)—\new{a 76\% reduction in $\tau$ over the strongest geometric baseline (GTS Exp.~2: $\tau=0.42$), the figure underlying the abstract claim.} ECLAIR and SD + LTS also perform well (0.14/0.38 and 0.17/0.35 for SD + LTS), confirming that structured document representations—semantic or learned—are beneficial.

{\paragraph*{Note on metric behaviour.} While $\tau$ degrades from Exp.~1 to Exp.~2 for all methods (as expected with added detection noise), $F$ often \textit{improves} slightly. This is a normalization artefact: missed detections in Exp.~2 reduce $n$, deflating the $\lfloor n^2/2 \rfloor$ denominator independently of ordering quality. We therefore rely primarily on $\tau$ for cross-experiment comparison.}

\begin{table}[!htbp]
\centering
\caption{Reading Order Accuracy – Task 2}
\begin{tabular}{lcccc}
\toprule
\textbf{Method} & \multicolumn{2}{c}{\textbf{Exp. 1}} & \multicolumn{2}{c}{\textbf{Exp. 2}} \\
\cmidrule(lr){2-3} \cmidrule(lr){4-5}
 & $\tau$ ↓ & $F$ ↓ & $\tau$ ↓ & $F$ ↓ \\
\midrule
SD + Generative LLM & 0.27 & 0.42 & 0.26 & 0.45 \\
\bottomrule
\end{tabular}
\label{tab:task2_results}
\end{table}

{
\subsection{LLM Comparison}

We swap the LLM in the pipeline across four models varying in size and deployment mode to assess the pipeline's sensitivity to the underlying LLM (Table~\ref{tab:llm_comparison}, Task~1 Exp.~1).
}

\begin{table}[!htbp]
\centering
\small
\caption{\new{Comparison of LLMs in the SD\,+\,Generative LLM pipeline (Task~1, Exp.~1). Models are grouped by deployment mode. Cost is estimated per 1,000 pages.}}
\begin{tabular}{llcccc}
\toprule
\textbf{Model} & \textbf{Mode} & \textbf{Params} & $\tau$ ↓ & $F$ ↓ & \textbf{Cost/1k p.} \\
\midrule
Ministral-3B               & Local   & 3B   & 0.12 & 0.28 & very low \\
Mistral-8B-Instruct-2410   & Local   & 8B   & 0.07 & 0.20 & low \\
\midrule
Mistral-Large              & API     & --   & 0.07 & 0.20 & high \\
Gemini~3.0~Flash             & API     & --   & 0.03 & 0.11 & high \\
\bottomrule
\end{tabular}
\label{tab:llm_comparison}
\end{table}

\new{This comparison addresses the \emph{bootstrapping} use case: small local models (Ministral-3B, Mistral-8B) offer the lowest cost for offline annotation, while cloud-API models (Gemini~3.0~Flash, Mistral-Large) trade cost for accuracy. Ministral-3B is clearly weaker than Mistral-8B; Mistral-Large performs on par with Mistral-8B, suggesting that scaling alone within the Mistral family is insufficient. Gemini outperforms all Mistral models, reaching the lowest scores; we cannot attribute the gap to a single cause (architecture, scale, or language coverage all plausibly contribute).}

W det.\,+\,LTS outperforms W ded.\,+\,LTS, confirming that explicit whitespace detection helps with nested column layouts; both still struggle on irregular structures. GTS underperforms layout-sensitive models, especially on column splits (0.42/0.50 in Exp.~2). \new{MLP scores near-randomly ($\tau \approx 0.70$): it lacks sequential inductive bias and is trained on fewer than 70 pages, insufficient for learning reliable spatial layouts.}

\paragraph*{Generative LLM alone: a systematic failure.} The standalone Generative LLM scores $\tau=0.75$ / $F=0.82$ in Exp.~1—well above the random baseline ($\tau\approx 0.5$): the model is not confused but \textit{systematically inverting} the reading order. We hypothesize two compounding factors: (1) Western Armenian is sparsely covered by Mistral's pretraining; (2) without spatial cues, the model reorders text thematically rather than spatially, conflicting with column structure. Substituting Gemini~3.0~Flash for Mistral-8B within the same SD\,+\,LLM pipeline lowers $\tau$ from $0.07$ to $0.03$ (Table~\ref{tab:llm_comparison}); we cannot disentangle architecture, scale, and language coverage from a single substitution, but the gap is consistent with stronger underlying capabilities on this task.

For Task 2 (Table~\ref{tab:task2_results}), \new{given the large gap in Task 1, we evaluate only SD + Generative LLM since other methods are unlikely to handle inter-page dependencies.} It remains effective: $\tau=0.27$/$F=0.42$ (Exp.~1) and $\tau=0.26$/$F=0.45$ (Exp.~2). The higher scores than in Task~1 ($\tau=0.07$) reflect the difficulty of long-range, cross-page dependencies. \new{Notably $F$ worsens but $\tau$ marginally improves between experiments—the opposite of Task~1—indicating that cross-page errors stem more from structural ambiguity than from local displacement.} Empirically, $\tau<0.1$ corresponds to minor line-level swaps, whereas $\tau>0.3$ signals disruption across columns or content blocks; SD + Generative LLM in Task~1 falls in the former regime.

{\paragraph*{OCR error cascade.} The $\tau$ degradation from Exp.~1 to Exp.~2 reflects both detection noise and OCR errors. SD\,+\,Generative LLM shows the smallest relative degradation ($0.07 \to 0.10$, +43\%) versus W det.\,+\,LTS ($+50\%$) and GTS ($+20\%$): the LLM can leverage partial contextual cues even when individual transcriptions are noisy.}

{\paragraph*{MLP anomaly.} For MLP, $\tau > F$ in both experiments ($0.70 > 0.68$; $0.73 > 0.66$). Although $F \geq \tau$ holds asymptotically, the normalized denominators $n(n-1)/2$ vs $\lfloor n^2/2 \rfloor$ can invert this for small $n$ ($n\leq5$), suggesting some test pages contain very few detected blocks.}

\section{Conclusion}

We benchmarked reading order strategies for historical Armenian newspapers and introduced a dedicated annotated dataset. Combining semantic zone detection with a generative LLM yields the most accurate and robust predictions in both single- and multi-page settings; we frame this approach as a \emph{data bootstrapping} tool—generating candidate orders that humans verify—rather than a production system. Our LLM comparison (Ministral-3B, Mistral-8B, Mistral-Large, Gemini~3.0~Flash) informs the quality/cost trade-off for this role.

\paragraph*{Limitations and future work.} On simple, unambiguous layouts, classical geometric methods (GTS, W det.\,+\,LTS) suffice and the LLM cost is unjustified; the proposed pipeline shines when the layout itself is ambiguous. Three directions follow. (i) Reduce the quadratic LLM-call count via smarter pairwise selection (transitive pruning, tournament sort, geometry-based filtering). (ii) Improve Armenian—especially Western Armenian—language coverage in the underlying LLMs, a plausible contributor to the Gemini gap and a long-term need for the field; this calls for larger corpora and dedicated language models. (iii) Use the bootstrapped annotations to train layout-aware models (LayoutLM, LayoutReader) for Armenian, removing the LLM from inference. Extending the approach to other endangered scripts and to more complex multi-article layouts are concrete next steps. Finally, we did not evaluate a full vision-language approach in which a VLM directly ingests the page image and returns the ordered text—an avenue that could in principle bypass both the OCR and the explicit SD detection stages, and worth exploring as VLM coverage of low-resource scripts improves. In any case, the bootstrapped annotations produced by our pipeline provide the supervision needed to fine-tune compact local models (e.g., Qwen3-VL-3B/8B) toward an offline, deployment-ready alternative.

\paragraph*{Acknowledgements, disclosure, data.} This research received support from ANR DALiH (ANR-21-CE38-0006); Claude AI was used to help condense the paper and proofread the English (any remaining mistakes are our own). The authors declare no competing interests. OCR model: \url{https://github.com/calfa-co/hye-tesseract}; dataset and code: \url{https://github.com/CVidalG/HYEpress-RO}.

{\footnotesize
\bibliographystyle{splncs04}
\bibliography{references}
}

\clearpage

\end{document}